\documentclass[conference]{IEEEtran}
\IEEEoverridecommandlockouts
\usepackage{cite}
\usepackage{hyperref}
\usepackage{amsmath,amssymb,amsfonts}
\usepackage{graphicx}
\usepackage{algorithm}
\usepackage{algpseudocode}
\usepackage{multirow}
\usepackage{adjustbox}
\algrenewcommand{\algorithmicrequire}{\textbf{Input:}}
\algrenewcommand{\algorithmicensure}{\textbf{Output:}}

\usepackage{textcomp}
\usepackage{xcolor}
\usepackage{pgfplots}
\pgfplotsset{compat=1.18}
\usepackage{tikz}
\usepackage{subcaption}
\usepackage{booktabs}
\hypersetup{hidelinks}
\usepackage{orcidlink}

\begin{document}
\title{GazeVLM: Gaze-Guided Foveated Inference for Efficient Vision-Language Models}

\author{
\IEEEauthorblockN{Lei Wang, Jiawen Qi, Qinyu Chen\orcidlink{0009-0005-9480-6164},~\IEEEmembership{Senior Member,~IEEE}}

\thanks{*Lei Wang and Jiawen Qi equally contributed to the work.}
\thanks{Lei Wang, Jiawen Qi and Qinyu Chen are with Leiden Institute of Advanced Computer Science (LIACS), Leiden University, 2300 RA Leiden, The Netherlands. (q.chen@liacs.leidenuniv.nl)}

}

\maketitle

\begin{abstract}
Vision-language models (VLMs) achieve remarkable multimodal performance but introduce substantial computational cost due to the large number of visual tokens. Existing approaches typically rely on token pruning, sparse attention, or additional control modules, often introducing accuracy--efficiency trade-offs.
This paper presents GazeVLM, a training-free framework that leverages human gaze as a natural guidance signal for efficient inference. GazeVLM allocates computation to gaze-relevant regions through a gaze-guided region-of-interest (ROI) representation and an optional two-scale design that preserves global context. 
We also construct GazeVLM-VQA, to the best of our knowledge, the first real-world wearable-gaze visual question answering (VQA) dataset specifically constructed for VLM inference, containing reference images, human gaze traces, and gaze-aligned question--answer pairs.
Experiments on the GazeVLM-VQA benchmark demonstrate that GazeVLM improves both answer quality and efficiency across Qwen2.5-VL-3B/7B and Qwen3-VL-8B models. 
It reduces visual tokens by up to 84.2\%, prefill FLOPs by up to 92.8\%, and achieves up to 11.6$\times$ prefill speedup, while improving answer-match accuracy by up to 8.6 percentage points (pp). 
These results show that aligning VLM computation with human visual attention provides a promising path toward efficient VLM inference.
We publicly release the
\href{https://drive.google.com/drive/folders/1mdi-GezeIkmege3segUQCnvnK6P-skaw?usp=sharing}{GazeVLM-VQA dataset}.
\end{abstract}

\section{Introduction}
Vision-Language Models (VLMs) demonstrate remarkable capabilities in tasks from basic image captioning~\cite{bai2025qwen2} to advanced visual question answering~\cite{hartsock2024vision}. 
They are increasingly being considered for interactive applications on augmented- and virtual-reality (AR/VR) devices~\cite{meta2024llama}. However, high-resolution visual inputs are represented as hundreds or thousands of visual tokens, making visual prefill a major source of latency and computation.
This is inefficient for task-oriented interaction, where only a small fraction of the scene may be relevant to the user’s current focus.

Deploying VLMs on such resource-constrained AR/VR devices is challenging.
Conventional VLMs allocate computation broadly across the visual input, even though many visual tokens may correspond to regions that are irrelevant to the current task.
High-resolution dense visual inputs yield long visual-token sequences, increasing prefill computation and latency. This overhead is particularly undesirable on battery-constrained AR/VR devices, motivating input representations that prioritize task-relevant regions.

Interestingly, many recent AR/VR devices~\cite{Varjo2025,apple2024visionpro,pico2024neo3pro} are already equipped with, or are moving toward integrating, eye-tracking modules~\cite{GazeHTA2025icra,facet2025icra,tan2025,3et2023}, which are primarily used for intuitive interaction and gaining insights into users’ attention. 
This provides a unique opportunity to exploit human physiological signals for efficient inference of VLMs. Human gaze is a direct behavioral manifestation of visual attention, reflecting perceptual and cognitive processes that determine which regions of a scene are most relevant for the task. Unlike artificial saliency models~\cite{han2025adafv}, gaze encodes real-time, user-specific intent and thus offers a natural guidance signal for multimodal inference.

In this work, we propose \textbf{GazeVLM}, a training-free foveated inference framework that leverages measured gaze traces as a task-dependent signal for efficient VLM inference. 
Our main contributions are summarized as follows:
\begin{enumerate}
    \item 
    We introduce GazeVLM, a training-free and plug-and-play framework that uses measured human gaze to identify task-relevant visual regions. GazeVLM processes a high-resolution gaze-guided ROI and optionally retains a low-resolution global view, reducing redundant visual tokens without modifying the VLM.

    \item 
    We construct GazeVLM-VQA, comprising 268 real-world images, 718 question--answer pairs, and wearable eye-tracking traces collected across 12 real-world scene categories.
    To the best of our knowledge, it is the first real-world wearable-gaze VQA dataset. 

    \item 
    We evaluate GazeVLM on GazeVLM-VQA across Qwen2.5-VL-3B, Qwen2.5-VL-7B, and Qwen3-VL-8B, and further conduct cross-dataset evaluation on VOILA-COCO. On GazeVLM-VQA, GazeVLM reduces visual-token usage by up to 84.2\%, and prefill latency by up to 11.6$\times$, while improving answer-match accuracy by up to 8.6 percentage points (pp).
\end{enumerate}

\section{GazeVLM-VQA Dataset}
\label{sec:dataset}
\subsection{Dataset overview}
Existing gaze-aligned VQA datasets are primarily constructed from laboratory data or existing image corpora. To evaluate gaze-guided VLM inference under a realistic sensing setting, we collect GazeVLM-VQA using a Pupil Labs Neon wearable eye tracker. 
GazeVLM-VQA contains 268 reference images at $4032\times3024$ resolution, 718 question--answer pairs, and spatially aligned gaze traces. It covers 12 scene categories across indoor public spaces (library, museums, etc.), indoor commercial spaces (supermarkets, clothing stores, etc.), and outdoor environments (parks, architecture, etc.), including text-rich and fine-grained scenes.

\subsection{Question--gaze pair construction}
During scene observation, the participant raised questions about
attended objects, attributes, text, and scene relations. 
In the meantime, these questions were recorded, allowing them to reflect the participant's immediate information needs. 
For each question–answer pair, we retained an approximately one-second gaze segment corresponding to the period in which the participant attended to the visual evidence relevant to that question.
The reference answer was then verified from the corresponding high-resolution image.
The questions are categorized into five types according to
their primary reasoning requirement: spatial reasoning, attribute
recognition, OCR/text recognition, existence, and semantic reasoning. Table~\ref{tab:question-distribution} summarizes the resulting distribution.

\begin{table}[t]
  \caption{Distribution and representative question examples in GazeVLM-VQA.}
  \label{tab:question-distribution}
  \centering
  \setlength{\tabcolsep}{4pt}
  \renewcommand{\arraystretch}{0.9}
  \resizebox{\columnwidth}{!}{%
    \begin{tabular}{@{}l c l@{}}
      \toprule
      Question type & Count (\%) & Example question \\
      \midrule
      Spatial Reasoning
        & 228 (31.75\%)
        & \begin{tabular}[c]{@{}l@{}}What is the name of the album to the right\\of the Bruno Mars album?\end{tabular} \\
      Attribute Recognition
        & 171 (23.82\%)
        & What is the material of the stool legs? \\
      OCR/Text Recognition
        & 164 (22.84\%)
        & When was the Cretaceous period? \\
      Existence
        & 109 (15.18\%)
        & \begin{tabular}[c]{@{}l@{}}Is there a computer mouse visible on\\the desk?\end{tabular} \\
      Semantic Reasoning
        & 46 (6.41\%)
        & \begin{tabular}[c]{@{}l@{}}What type of room is this most likely\\to be?\end{tabular} \\
      \midrule
      \textbf{Total}
        & \textbf{718 (100.00\%)}
        & -- \\
      \bottomrule
    \end{tabular}%
  }
\end{table}

\section{GazeVLM Framework}
\label{sec:Method}

\begin{figure}[t]
\begin{center}
\includegraphics[width=1\linewidth]{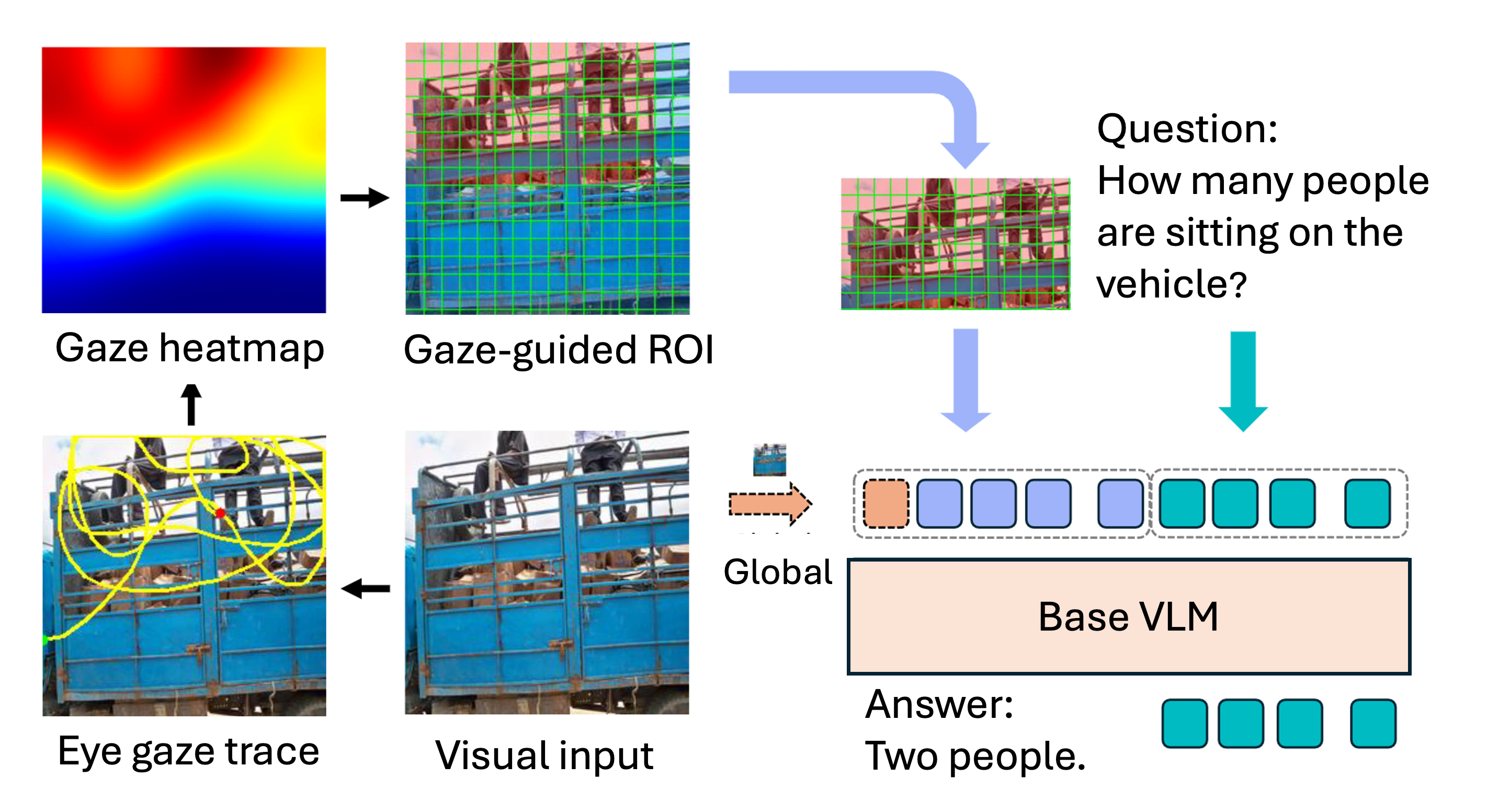}
\end{center}
\caption{
Overview of GazeVLM.
}
\label{fig:overview}
\end{figure}

Based on the aforementioned insights, we propose GazeVLM, a training-free gaze-guided foveated inference framework (Fig.~\ref{fig:overview}).
It supports two input configurations: a gaze-guided ROI (Gaze-ROI) and a two-scale representation that additionally preserves
low-resolution global context (Gaze-2SCALE).

\subsection{Gaze-guided ROI Extraction}

Let the input image be $I \in \mathbb{R}^{H \times W \times 3}$ 
and its corresponding gaze heatmap
\begin{equation}
G \in [0,1]^{H \times W},
\qquad
\sum_{x,y} G(x,y)=1.
\end{equation}
The raw gaze traces are projected onto the image plane and smoothed
with a Gaussian kernel to obtain a continuous attention map.
Higher values of $G(x,y)$ indicate greater gaze mass around pixel $(x,y)$, corresponding to more gaze-concentrated regions. 
This produces a smooth heatmap that reflects the distribution of visual attention across the image.

To extract a compact ROI, we rank all pixels in descending order
of gaze mass. Let $\pi_1,\ldots,\pi_{HW}$ denote the resulting
ordering, such that
$G(\pi_1)\geq \cdots \geq G(\pi_{HW})$.
We retain the smallest prefix whose cumulative gaze mass reaches
$\rho$:
\begin{equation}
\mathcal{S}_{\rho}
=
\{\pi_1,\ldots,\pi_{k^\star}\},
\qquad
k^\star
=
\min\left\{
k:\sum_{j=1}^{k}G(\pi_j)\geq\rho
\right\}.
\label{eq:gaze_mass}
\end{equation}
The axis-aligned bounding box enclosing
$\mathcal{S}_{\rho}$ defines the ROI
$B=(x_0,y_0,x_1,y_1)$, from which we obtain the crop
$I_r=I[B]$.

\subsection{Gaze-2SCALE Input Representation}
Humans do not process the entire visual field at uniform resolution. 
Instead, the fovea provides high-resolution vision at the attended region,  while the periphery captures the broader scene at lower resolution. 

Once the ROI is extracted, the next question is: \emph{should we only feed the ROI to the model}?  
While the ROI contains the most task-relevant details, discarding the full image entirely would remove global cues such as scene layout, object co-occurrence, or background context, which may still be essential for correct reasoning.  
On the other hand, if we only feed the full image at a fixed resolution, the ROI can only be represented at low fidelity, making it difficult for the model to capture fine-grained details (e.g., small objects or text).  

To resolve this trade-off, we introduce a Gaze-2SCALE input inspired by the fovea-periphery organization of the human visual system.  
Specifically, we construct two complementary image views:
\begin{itemize}
    \item Global view ($I_g$): the original image is downsampled to a coarse resolution $H_g \times W_g$.  
    This preserves the holistic scene layout at a low computational cost, analogous to peripheral vision.
    \item ROI view ($I_r$): the gaze-driven crop with resolution $H_r \times W_r$.  
    This provides fine-grained detail in the attended region, analogous to foveal vision.
\end{itemize}

Both $I_g$ and $I_r$ are processed by the encoder to generate visual tokens, which are then concatenated with the text query $T$ to form the final model input:
\begin{equation}
X = \mathcal{P}\big( \{I_g, I_r\}, \, T \big),
\end{equation}
where $\mathcal{P}$ denotes the process that aligns visual and textual modalities.

Overall, the Gaze-2SCALE representation mirrors human perception: the global view maintains peripheral awareness, while the ROI view supplies foveated detail.  
This design enables the model to allocate its limited token budget more effectively, improving both efficiency (fewer irrelevant tokens) and accuracy (sharper focus on gaze-attended content).

\section{Experiments}

\subsection{Experimental Setup}

We evaluate GazeVLM on GazeVLM-VQA using
Qwen2.5-VL-3B/7B~\cite{bai2025qwen2}
and Qwen3-VL-8B~\cite{qwen3vltech},
and on VOILA-COCO~\cite{yan2024voila}
using Qwen2.5-VL-3B/7B.
VOILA-COCO comprises approximately 20k images
and 70k QA pairs with mouse-based gaze proxies.
Experiments are conducted on an NVIDIA
A100-SXM4-40GB GPU.
Answer quality is evaluated by normalized exact-match
accuracy on GazeVLM-VQA and GPT-4o-based scoring
on VOILA-COCO.
The latter uses order-swapped pairwise comparisons
and a weighted score combining coverage (40\%),
accuracy (40\%), detail (15\%), and fluency (5\%).
Efficiency is measured by visual/total token counts,
prefill FLOPs, and prefill latency under identical
model and hardware configurations.

\subsection{Main Quality--Efficiency Trade-off}

\begin{table}[t]
\centering
\caption{Quality--efficiency results on GazeVLM-VQA ($\rho=0.5$).}
\label{tab:main_results}

\small
\setlength{\tabcolsep}{1.5pt}
\renewcommand{\arraystretch}{0.8}

\begin{adjustbox}{max width=\columnwidth}
\begin{tabular}{llccccc}
\toprule
Model & Mode &
\shortstack{Acc.\\(\%) $\uparrow$} &
\shortstack{Visual\\Tok. $\downarrow$} &
\shortstack{Total\\Tok. $\downarrow$} &
\shortstack{FLOPs\\(T) $\downarrow$} &
\shortstack{Latency (s) $\downarrow$\\Speedup $\uparrow$} \\
\midrule

\multirow{3}{*}{Qwen2.5-VL-3B}
& BASE         & 73.54          & 15.55k & 15.59k & 106.5 & 3.20/1.0$\times$ \\
& Gaze-ROI     & 82.03          & 2.46k  & 2.50k  & 7.7   & \textbf{0.45/7.1$\times$} \\
& Gaze-2SCALE  & \textbf{82.17} & 2.46k  & 2.50k  & 7.7   & 0.46/7.0$\times$ \\
\midrule

\multirow{3}{*}{Qwen2.5-VL-7B}
& BASE         & 75.63          & 15.55k & 15.59k & 179.6 & 3.76/1.0$\times$ \\
& Gaze-ROI     & \textbf{78.69} & 2.46k  & 2.50k  & 16.1  & 0.54/7.0$\times$ \\
& Gaze-2SCALE  & 77.99          & 2.46k  & 2.50k  & 16.1  & 0.54/7.0$\times$ \\
\midrule

\multirow{3}{*}{Qwen3-VL-8B}
& BASE         & 80.78          & 11.84k         & 11.87k         & 156.1         & 3.60/1.0$\times$ \\
& Gaze-ROI     & 83.70          & \textbf{1.88k} & \textbf{1.91k} & \textbf{14.7} & \textbf{0.31/11.6$\times$} \\
& Gaze-2SCALE  & \textbf{83.98} & 1.94k          & 1.98k          & 15.1          & 0.32/11.3$\times$ \\
\bottomrule
\end{tabular}
\end{adjustbox}
\end{table}

Table~\ref{tab:main_results}
summarizes the quality--efficiency results on GazeVLM-VQA at $\rho=0.5$.
Across all evaluated Qwen models, gaze-guided inference substantially reduces computational cost while maintaining or improving answer quality.
Compared with full-image inference (BASE), ROI reduces visual-token counts by 84.1--84.2\% and estimated prefill FLOPs by 90.6--92.8\%.
These reductions translate into measured runtime speedups ranging from 7.0$\times$ to 11.6$\times$.
At the same time, answer-match accuracy improves from
73.54\% to 82.03\% on Qwen2.5-VL-3B, from 75.63\% to 78.69\% on Qwen2.5-VL-7B, and from 80.78\% to 83.70\% on Qwen3-VL-8B.
The Gaze-2SCALE configuration exhibits a similar computational profile while achieving the highest accuracy for Qwen2.5-VL-3B and Qwen3-VL-8B.
Overall, both gaze-guided configurations consistently move the operating point toward a more favorable quality--efficiency trade-off.

\subsection{Effect of Gaze-Guided Region Selection}

\begin{figure}[t]
\begin{center}
\includegraphics[width=1\linewidth]{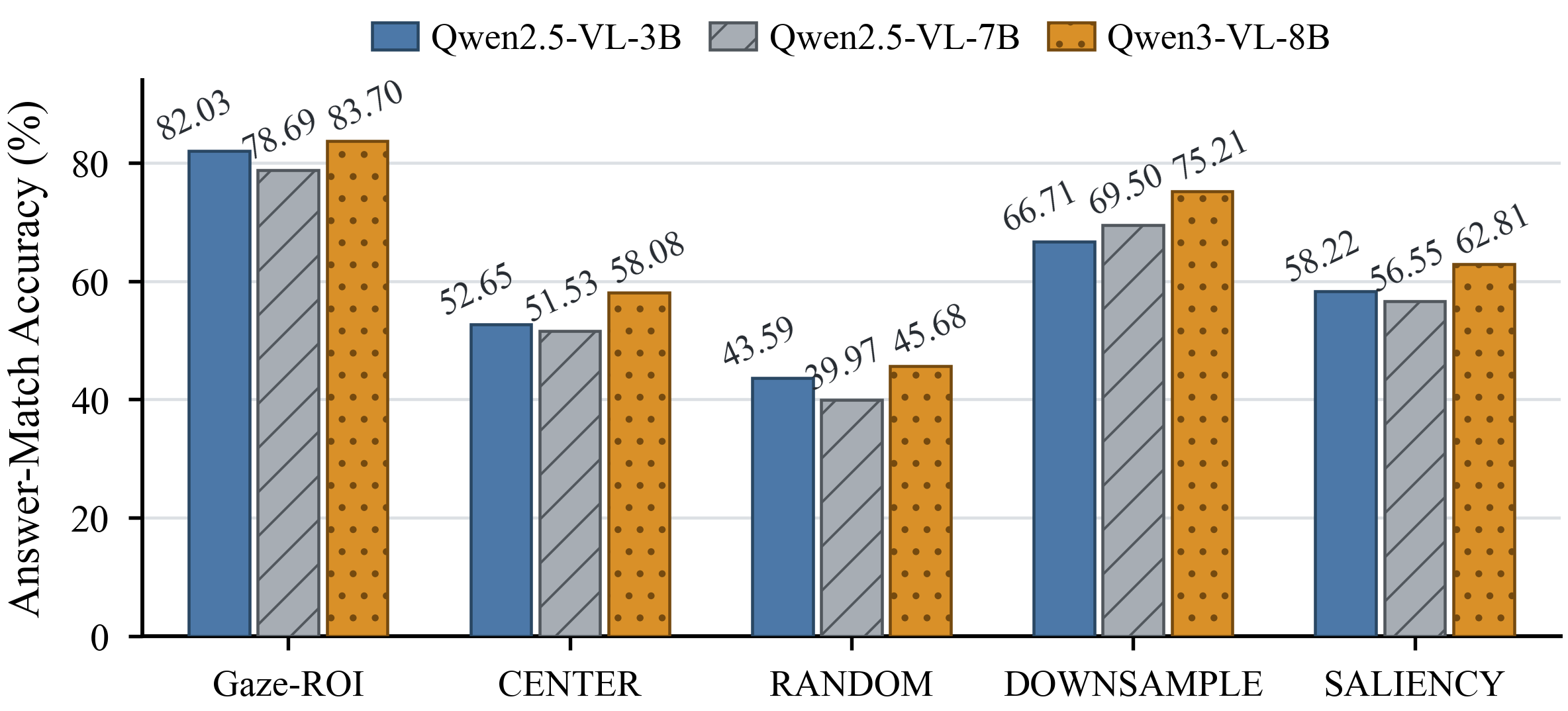}
\end{center}
\caption{Answer-match accuracy of Gaze-ROI and non-gaze
visual selection strategies (CENTER, RANDOM, DOWNSAMPLE, SALIENCY) under matched per-sample visual budgets.
}
\label{fig:nongaze}
\end{figure}

Fig.~\ref{fig:nongaze} compares Gaze-ROI with four
non-gaze strategies (CENTER, RANDOM, DOWNSAMPLE,
and SALIENCY~\cite{DeepGaze_IIE}) under matched per-sample visual budgets.
Gaze-ROI achieves the highest answer-match accuracy across all three
models. 
Although Saliency-ROI outperforms CENTER and RANDOM, it remains below DOWNSAMPLE for all three models. This indicates that task-agnostic saliency can identify visually prominent regions, but does not reliably localize the evidence required by a specific question.
Compared with DOWNSAMPLE, Gaze-ROI improves accuracy by 15.32, 9.19, and 8.49 pp across the three models.
These results show that the gains arise not merely from cropping, matching the visual budget, or selecting salient image regions, but
from using question-aligned gaze to allocate the available visual budget to task-relevant regions.

\begin{figure}[t]
\begin{center}
\includegraphics[width=0.9\linewidth]{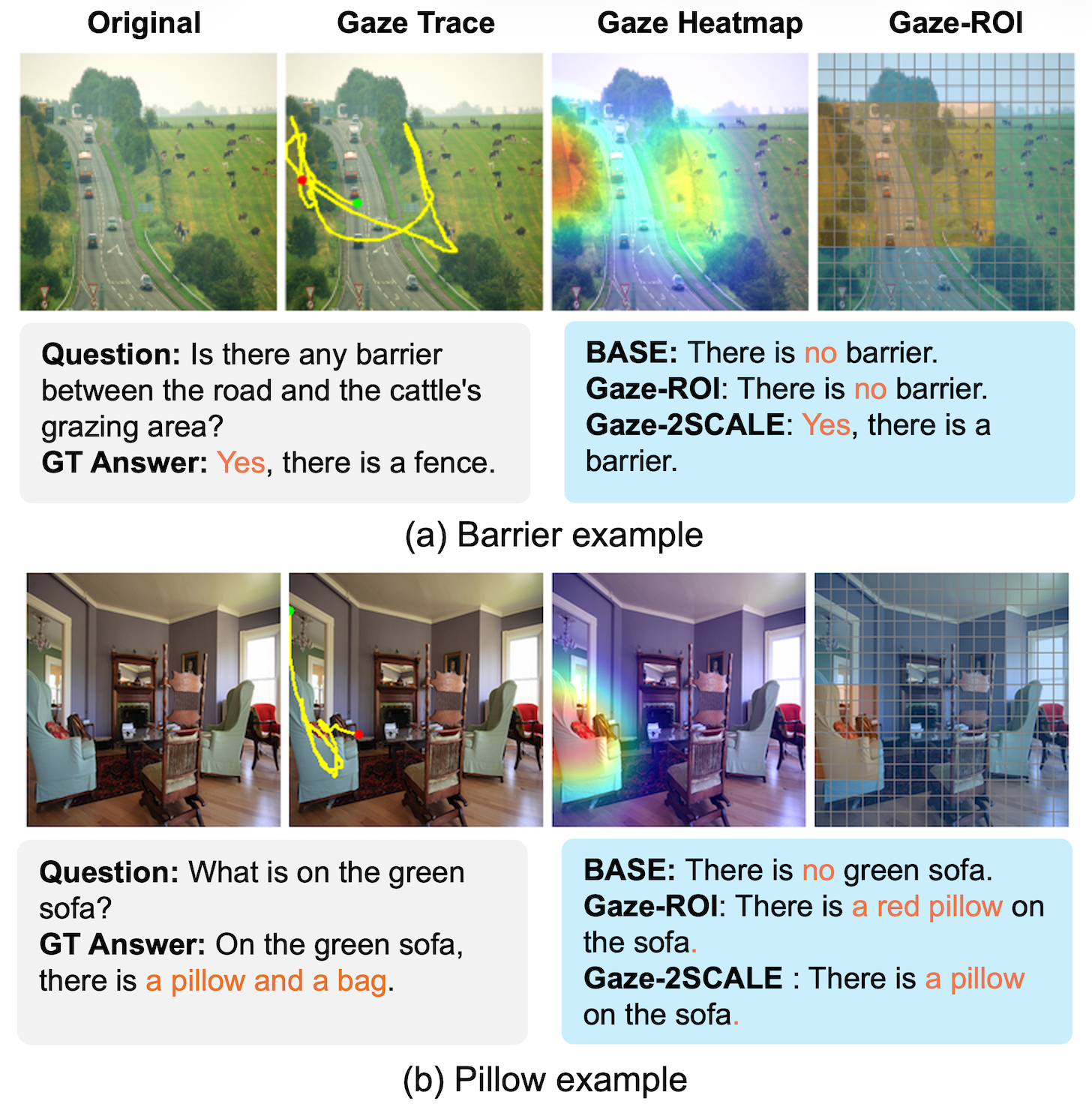}
\end{center}
\caption{
Qualitative results (Qwen2.5-VL-3B, $\rho=0.3$).}
\label{fig:qualitative_twoscale}
\end{figure}

\subsection{Does Global Context Provide Benefits?}

\begin{figure*}[ht]
\begin{center}
\includegraphics[width=1\linewidth]{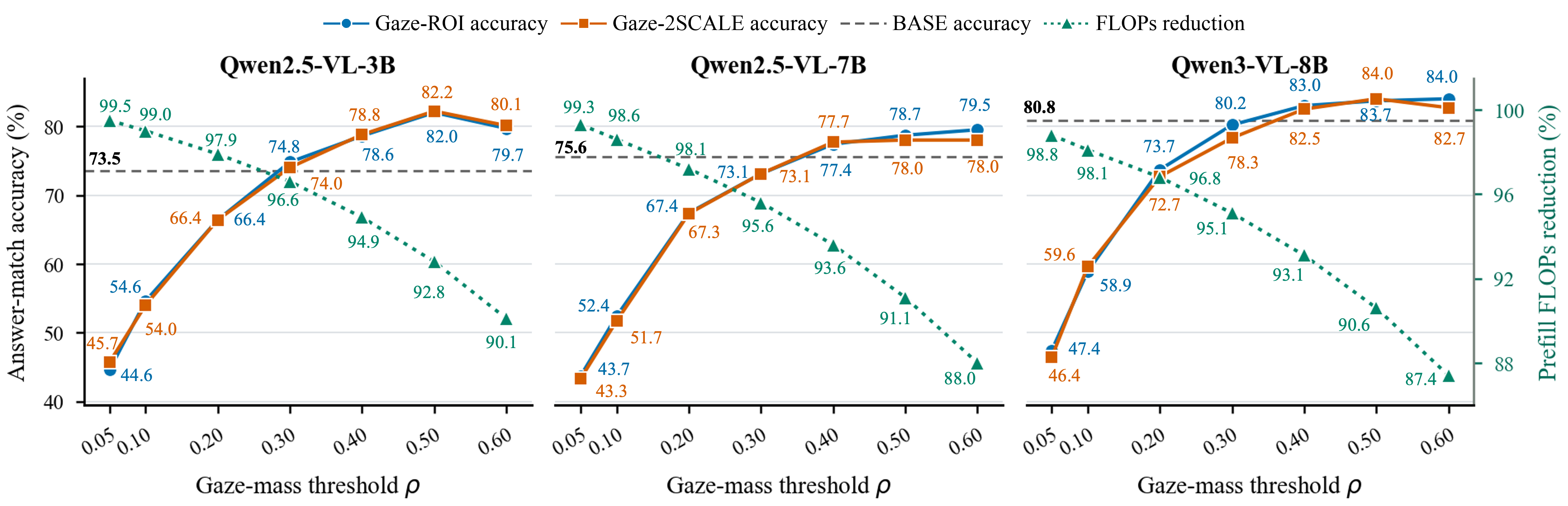}
\end{center}
\caption{
Sensitivity to the gaze-mass threshold $\rho$ on GazeVLM-VQA.}
\label{fig:threshold_sensitivity}
\end{figure*}

As shown in Table~\ref{tab:main_results}, the performance difference between Gaze-ROI and Gaze-2SCALE inference is relatively small across all evaluated models. Qwen2.5-VL-3B and Qwen3-VL-8B obtain modest gains of 0.14 pp and 0.28 pp, respectively, whereas Qwen2.5-VL-7B exhibits a slight decrease of 0.70 pp.
The additional global image introduces only minimal computational overhead.
Compared with Gaze-ROI, the Gaze-2SCALE configuration increases runtime by at most 0.01\,s while preserving nearly all efficiency benefits of gaze-guided processing.
Fig.~\ref{fig:qualitative_twoscale} illustrates that
the global view can recover missing spatial context
when the ROI alone is insufficient, while providing
little additional benefit when the relevant evidence
is already captured by the gaze-guided crop.

\subsection{Generalization and Robustness}
\subsubsection{Cross-Dataset Validation on VOILA-COCO}

\begin{table}[t]
\centering
\caption{Validation on VOILA-COCO ($\rho=0.3$).}
\label{tab:voila_generalization}

\small
\setlength{\tabcolsep}{3pt}
\renewcommand{\arraystretch}{0.85}

\begin{adjustbox}{max width=\columnwidth}
\begin{tabular}{llcccc}
\toprule
Model & Mode & 
\shortstack{GPT-4o\\Score $\uparrow$} &
\shortstack{Visual\\Tok. $\downarrow$} &
\shortstack{Total\\Tok. $\downarrow$} &
\shortstack{FLOPs\\(G) $\downarrow$} \\
\midrule

\multirow{3}{*}{Qwen2.5-VL-3B}
& BASE        & 3.98          & 64.0          & 100.0         & 267.6 \\
& Gaze-ROI    & 4.45          & \textbf{15.6} & \textbf{51.8} & \textbf{158.6} \\
& Gaze-2SCALE & \textbf{4.51} & 19.0          & 62.6          & 186.9 \\
\midrule

\multirow{3}{*}{Qwen2.5-VL-7B}
& BASE        & 5.73          & 64.0          & 100.0         & 631.0 \\
& Gaze-ROI    & 5.84          & \textbf{15.6} & \textbf{51.8} & \textbf{374.0} \\
& Gaze-2SCALE & \textbf{5.95} & 19.0          & 62.6          & 433.2 \\
\bottomrule
\end{tabular}
\end{adjustbox}
\end{table}

We additionally evaluate GazeVLM on the VOILA-COCO dataset at $\rho=0.3$.
As shown in Table~\ref{tab:voila_generalization}, the trends observed on GazeVLM-VQA remain consistent.
For both Qwen2.5-VL-3B and Qwen2.5-VL-7B, ROI-based inference substantially reduces visual-token usage (from 64 to 15.6 tokens) and total context length (from 100 to 51.8 tokens), while simultaneously improving GPT-4o evaluation scores.
Furthermore, incorporating global context through the
Gaze-2SCALE design yields additional gains.
Compared with Gaze-ROI inference, Gaze-2SCALE improves the GPT-4o score from 4.45 to 4.51 on Qwen2.5-VL-3B and from 5.84 to 5.95 on Qwen2.5-VL-7B, while still reducing FLOPs by 30.2\% and 31.3\% relative to the full-image baseline.
These results demonstrate that the effectiveness of gaze-guided token allocation generalizes across different gaze collection protocols and VQA tasks.

\subsubsection{Sensitivity to the Gaze-Mass Threshold}

We analyze the sensitivity of GazeVLM to the retained gaze-mass threshold $\rho$ on GazeVLM-VQA. 
A larger $\rho$ retains a greater fraction of the gaze heatmap, thereby enlarging the selected ROI and preserving more visual context, but also increasing the number of visual tokens and the associated prefill computation.

As shown in Fig.~\ref{fig:threshold_sensitivity}, answer quality
improves rapidly for all the three models as $\rho$ increases from 0.05 to 0.30.
For Gaze-ROI inference, accuracy increases from 44.6\% to
74.8\% on Qwen2.5-VL-3B, from 43.7\% to 73.1\% on
Qwen2.5-VL-7B, and from 47.4\% to 80.2\% on Qwen3-VL-8B.
The Gaze-2SCALE configuration follows a similar trend, increasing from
45.7\% to 74.0\%, from 43.3\% to 73.1\%, and from 46.4\% to
78.3\%, respectively. 
At $\rho=0.30$, these results match or
approach the corresponding full-image baselines of 73.5\%, 75.6\%,
and 80.8\%, while ROI inference still reduces estimated prefill
FLOPs by 96.6\%, 95.6\%, and 95.1\%.

Beyond $\rho=0.30$, the quality gains begin to
saturate while computational savings continue to decrease.
Increasing $\rho$ from 0.30 to 0.40 still improves Gaze-ROI accuracy by 3.8, 4.3, and 2.8 pp for Qwen2.5-VL-3B,
Qwen2.5-VL-7B, and Qwen3-VL-8B, respectively. Further increasing
$\rho$ produces progressively smaller and less consistent gains.
At $\rho=0.60$, Qwen2.5-VL-3B declines under both Gaze-ROI and Gaze-2SCALE, Qwen2.5-VL-7B improves only slightly under Gaze-ROI, and Qwen3-VL-8B shows only a marginal Gaze-ROI gain while its Gaze-2SCALE accuracy decreases.
At large $\rho$, low-density or spatially distant gaze samples may
enlarge the ROI and introduce irrelevant background, diluting the
task-relevant evidence and making visual selection more difficult,
particularly for smaller models.
Overall, answer quality largely saturates around $\rho=0.40$--$0.50$, whereas computational cost continues to grow.

\section{Conclusion}
In this paper, we presented GazeVLM, a training-free framework that uses measured human gaze to guide efficient VLM inference. 
We also introduced GazeVLM-VQA, a real-world wearable-gaze benchmark. 
Across three Qwen-VL models, GazeVLM reduces visual-token usage by up to 84.2\% and achieves up to 11.6$\times$ prefill speedup, while improving answer-match accuracy by up to 8.6 pp. Cross-dataset evaluation on VOILA-COCO further demonstrates the generalizability of gaze-guided visual-token allocation.

\bibliographystyle{IEEEtran}
\bibliography{references}
\end{document}